%% file: main.tex
\RequirePackage{fix-cm}

\documentclass{article}      
\usepackage{mathtools}
\usepackage{multirow}
\usepackage{graphicx}
\usepackage{subfigure}
\usepackage{wrapfig}
\usepackage{url}
\RequirePackage[bookmarks,
                urlcolor=black,
                citecolor=black,
                linkcolor=black,
                pagecolor=black,
                colorlinks,
                hyperfigures]{hyperref}
\usepackage{csquotes}
\usepackage{todonotes}
\usepackage[noend,linesnumbered]{algorithm2e}
\usepackage{placeins}
\usepackage[inline]{enumitem}
\usepackage{array}
\usepackage{listings}
\usepackage{pgfplots}
\usepackage{pgfplotstable}
\usepackage[gen]{eurosym}
\usepackage{amsfonts}
\usepackage{amsmath}

\newcommand*\circled[1]{\tikz[baseline=(char.base)]{\node[shape=circle,draw,inner sep=0pt] (char) {#1};}}

\begin{document}

\title{The Role of Time and Data: Online Conformance Checking in the Manufacturing Domain}

\author{Florian Stertz$^1$, Juergen Mangler$^2$, Stefanie Rinderle-Ma$^2$\\
$^1$Faculty of Computer Science, \\ University of Vienna, Austria, \\ florian.stertz@univie.ac.at\\           
$^2$Department of Informatics, Technical University of Munich, \\ Germany, \{juergen.mangler,stefanie.rinderle-ma\}@tum.de}

\date{}

\maketitle

\begin{abstract} 
Process mining has matured as analysis instrument for process-oriented data in recent years.
Manufacturing is a challenging domain that craves for process-oriented technologies
to address digitalization challenges. We found that process mining creates high
expectations, but its implementation and usage by manufacturing experts such as process supervisors 
and shopfloor workers remain unclear to a certain extent. 
Reason $\circled{1}$ is that even though manufacturing allows for
well-structured processes, the actual workflow is rarely captured in a process
model. Even if a model is available, a software for orchestrating and logging
the execution is often missing.  Reason $\circled{2}$ refers to the work
reality in manufacturing: a process instance is started by a shopfloor worker
who then turns to work on other things. Hence continuous monitoring of the
process instances does not happen, i.e., process monitoring is merely a
secondary task, and the shopfloor worker can only react to problems/errors that
have already occurred. $\circled{1}$ and $\circled{2}$ motivate the goals of
this study that is driven by Technical Action Research (TAR). Based on the
experimental artifact TIDATE --a lightweight process execution and mining
framework-- it is studied how the correct execution of process instances can be
ensured and how a data set suitable for process mining can be generated at
run time in a real-world setting.  Secondly, it is investigated whether and how
process mining supports domain experts during process monitoring as a secondary
task. The findings emphasize the importance of online conformance checking in
manufacturing and show how appropriate data sets can be identified and
generated.   \end{abstract}


\section{Introduction}
\label{sec:intro}
\input{intro.tex}

\section{Methodology}
\label{sec:meth}
\input{meth.tex}

\section{Research Execution}
\label{sec:eval}
\input{eval.tex}

\section{Data analysis}
\label{sec:knowledge}
\input{knowledge.tex}

\section{Related Work}
\label{sec:rel}
\input{rel.tex}

\section{Conclusion}
\label{sec:concl}
\input{concl.tex}

\section*{Acknowledgment} 
This work has been partly funded by the
Austrian Research Promotion Agency (FFG) via the ``Austrian Competence Center
for Digital Production'' (CDP) under the contract number 881843. This work has
been supported by the Pilot Factory Industry 4.0, Seestadtstrasse 27, Vienna,
Austria.

\bibliographystyle{splncs04}
\bibliography{bib}
\end{document}

%% file: intro.tex
Literature emphasizes the importance of process mining to provide the necessary transparency for pushing digital transformation in manufacturing \linebreak\hfill\cite{Rein2020}. A recent Gartner study \cite{Gartner2020} reports a steep increase in process mining use cases for digital transformation and process automation. A domain that is at the forefront for both use cases is \textsl{manufacturing}. It combines high demands on process transparency and digital transformation and it combines the physical world (e.g., sensors, machines), human work, and manufacturing systems.

Though some use cases for applying process mining in the manufacturing domain
have been presented, \cite{Rein2020}, it remains still unclear how process
mining can be actually implemented and used by manufacturing experts (i.e.,
\textsl{process supervisors} and \textsl{shopfloor workers}), particularly for
small and medium size enterprises. The reasons are as follows:

\begin{itemize}
\item[$\circled{1}$] Even though the tasks of shopfloor workers are clearly structured
and can be viewed as a business process, the operated machines often times lack an actor
orchestrating the planned tasks
\item[$\circled{2}$] The work reality in manufacturing is that a process
instance is started by a shopfloor worker who then turns to work on other
things. Hence a continuous monitoring of the process instance does not
happen--i.e., process monitoring is merely a secondary task--and the shopfloor
worker can only react to problems/errors that have already occurred.
\end{itemize}

One way to tackle $\circled{1}$ is to have experts create a process model and
execute it in their working shifts. Creating a process model reflects an
increased work effort by the experts, with no guarantee that, the process model
is correctly enacted afterwards without a software that is orchestrating the
process. In addition, process mining techniques need a data set to be applied
on. Even though many resources create a log for themselves, the logs of the
resources need to be collected, linked, and transformed to be suitable for
process mining techniques.  This is expensive and maximally invasive, because
the way how information on the execution of process instances for process
mining is stored, is in so called \textsl{process execution logs} (logs for
short)\footnote{In general, a log --usually defined in the XES format
((\url{www.xes-standard.org})-- contains information on all process instances
that have been executed, i.e., captures the actual behavior of the process. The
log consists of so-called traces, where a trace carries the data about a
process instance it the executed tasks. Each executed task is stored as an
event, usually containing information on the executing resource, a time stamp
and a position in the lifecycle of the task.}.
On top of that, in order to meet the full spectrum
of possible data sources in the manufacturing domain, the collection of process
execution data has to be augmented with the collection of sensor and machining
data \cite{DBLP:conf/bpm/StertzRM20}. 

Automatic process monitoring during runtime constitutes a promising remedy for $\circled{2}$ because in case of deviations or problems, the system could alarm the domain expert instantly. Basically, \textsl{conformance checking} \cite{DBLP:books/sp/CarmonaDSW18} as one of the process mining tasks aims at measuring how well the behavior of a process instances expressed by the corresponding logs fits the behavior
described by a process model. Conversely, conformance checking can be used to detect deviations in behavior. Precisely, conformance checking determines \textsl{alignments} between a log and a process model based on \textsl{moves} that reflect matching as well as non-matching behavior. Costs can be assigned to moves for non-matching behavior. Conformance checking based on logs is conducted ex post/offline. However, for supporting process monitoring as secondary task conformance checking must be applied during runtime/online for currently running process instances. Hence, an \textsl{event stream} is used instead of a log.

\begin{figure}[htb!]
     \centering
    \includegraphics[width=\textwidth]{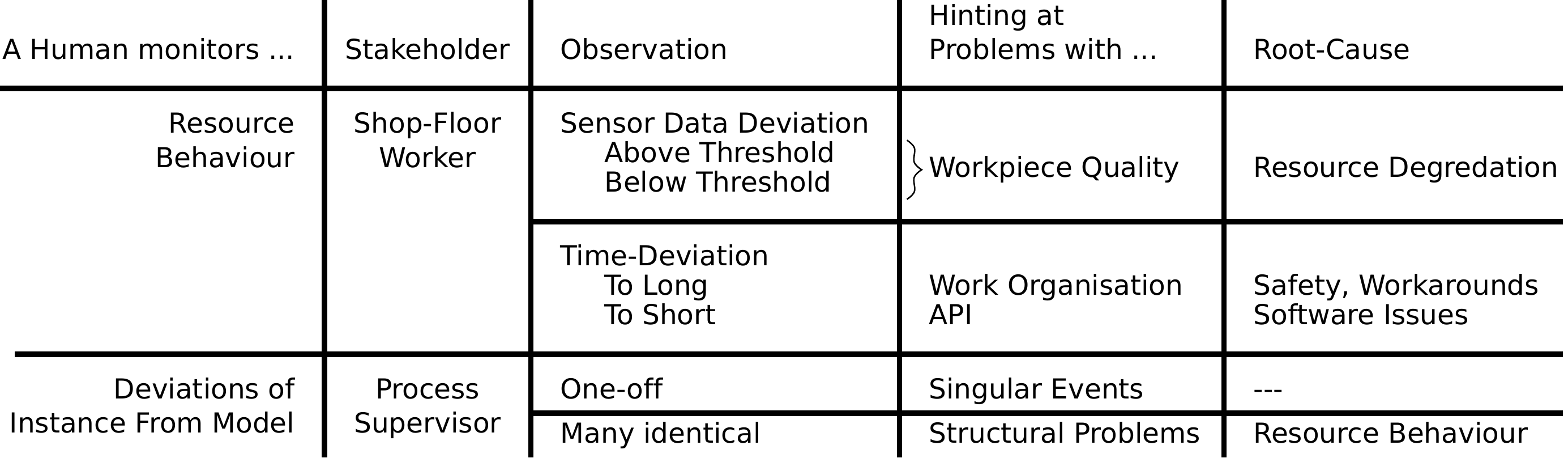}
    \caption{Time and data-aware conformance deviations and their root-causes in manufacturing}
    \label{fig:deviationsroot}
\end{figure}

For the above reasons \textsl{online conformance checking} is considered as suitable analysis technique for monitoring support in manufacturing. Fig. \ref{fig:deviationsroot} provides an overview what is exactly monitored by the shopfloor worker and the process supervisor. The shopfloor worker monitors the \textsl{resource behavior} based on sensor data and time. Deviations in the sensor data hint to problems with the workpiece quality and occur due to resource degradation \cite{DBLP:conf/bpm/StertzRM20}. Time-deviations are mostly caused by humans, e.g., someone stepping into the safety area of a machine causing a delay, and hint to problems with work organization. The process supervisor is responsible for monitoring deviations due to ad-hoc adaptations of instances--resolving singular events-- or to an evolution of the process model--resolving structural problems. In order to cover all these monitoring tasks, \textsl{data and time-aware online conformance checking} is required, targeting the question of \textsl{``what is the potential of using various types of data in process mining?''}$^2$.

$\circled{1}$ and $\circled{2}$ form the goals of this study. From a methodological point we employ Technical Action Research \cite{wieringa2012technical,wieringa2014design}. \textsl{''Technical action research (TAR) is the use of an experimental artifact to help a client and to learn about its effects in practice''} \cite{wieringa2014design}. Doing so, TAR constitutes the last step \textsl{``from the conditions of the laboratory to the unprotected conditions of practice''}. This study uses the lightweight process execution and mining framework TIDATE as experimental artifact. TIDATE consists of a lightweight and modular process engine\footnote{https://cpee.org/} for modeling and executing the manufacturing processes. Moreover, TIDATE features process logging mechanisms including the linking of sensor and other machining parameters to the logs. Finally, TIDATE features process mining techniques, particularly time and data-aware conformance checking during runtime \cite{DBLP:conf/bpm/StertzRM20,stertztemp}. TIDATE is applied in a manufacturing environment.

The findings are promising. From the different process mining tasks, process discovery seems to play a minor role in manufacturing whereas the importance of online conformance checking is emphasized. The role of time and data in conformance checking is emphasized, too, i.e., shopfloor workers and process supervisors are actually supported in their process monitoring task. The generation of data sets that are suitable for a product-oriented analysis is possible and does not conflict with the batch-oriented modeling of the processes. Linking sensor data requires a certain upfront effort.

The paper is structured as follows. The methodology
and setup of the study is explained in Sec. \ref{sec:meth}. The implemented approach is then
tested and evaluated in Sec. \ref{sec:eval}. Section \ref{sec:knowledge}
answers the knowledge questions of the approach and discusses further
improvements. Related work is presented in Sec.  \ref{sec:rel} and the paper is
concluded in Sec. \ref{sec:concl} with some finishing thoughts and information
on the planned future work.

%% file: meth.tex
This study employs Technical Action Research (TAR)
\cite{wieringa2012technical,wieringa2014design} aiming at the validation of an
artifact in a realistic case, i.e., the implementation in a real-world
organization. This is a major distinction to other forms of Action Research as
TAR is technology-driven instead of problem-driven. Section
\ref{sub:researchproblem} outlines the research context of this study, Sect.
\ref{sub:design} the research design, and Sect. \ref{sub:validation} the
artifact validation.

\subsection{Research Context}
\label{sub:researchproblem}

In TAR, the researcher takes three logically separate roles, i.e., \textbf{technical researcher}, \textbf{empirical researcher}, and \textbf{helper}. Consequently, the researcher in TAR, is allowed to guide domain
experts using the artifact, to interfere, and aims to answer research questions
with the results of the experiment.

\begin{figure}
  \centering
  \includegraphics[width=0.6\textwidth]{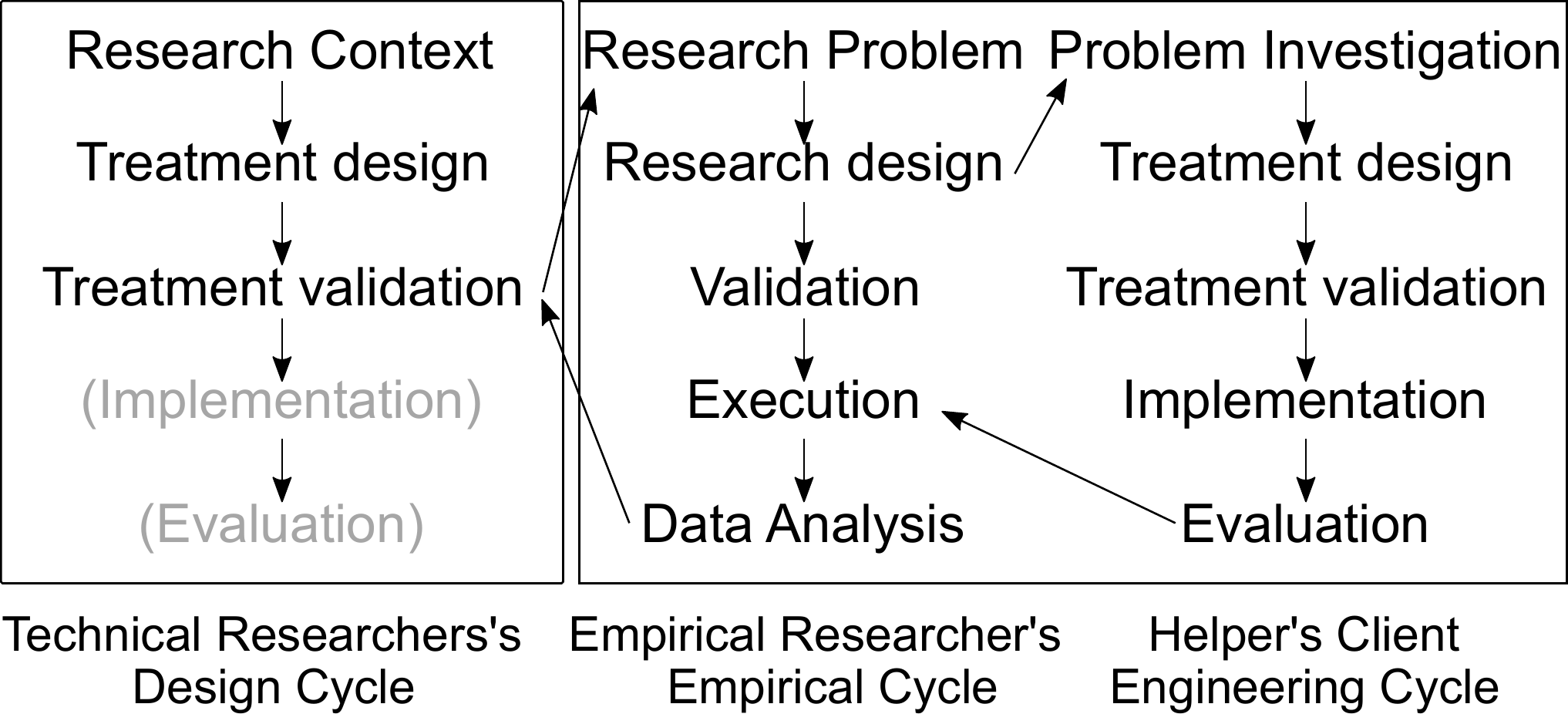}
  \caption{TAR cycles (based on \cite{wieringa2014design})}
  \label{fig:tar}
\end{figure}

Fig. \ref{fig:tar} (based on \cite{wieringa2012technical}) shows the three cycles used in TAR that reflect the three roles of the researcher in TAR.

The technical researcher's design cycle defines the research context of a study. According to \cite{wieringa2014design}, at first, the research problem is investigated and the \textbf{research context} is defined by describing the goals of the created artifact and the current knowledge of the environment.
For this study, the experimental artifact is the lightweight process execution
and mining framework TIDATE and the practical setting is a manufacturing
environment. The clients are the domain experts, i.e., the shopfloor workers
and the process supervisors. In this study, the clients can benefit from time and data-aware conformance checking techniques
by detecting errors in the behavior of the manufacturing processes as soon as possible. Moreover, the clients benefit from a minimally invasive generation of product-oriented logs/event streams based on their batch-oriented process models.
In the following, we define the research context by dividing the goals into knowledge and improvement goals.

\begin{description}
  \item[\textbf{Knowledge goal:}] Can TIDATE help shopfloor workers and process supervisors in a manufacturing environment
  in a useful way?
  \item[\textbf{Improvement goals:}] \ 
  \begin{itemize}
    \item Generate event streams for time and data-aware conformance checking during runtime in a product-oriented manner.
    \item Highlight time deviations in process instances as soon as possible.
    \item Highlight data deviations linked to process instances as soon as possible.
    \item Highlight deviating behavior in the execution order of events in process instances as soon as possible.
  \end{itemize}
  \item[\textbf{Current Knowledge}:] Process mining techniques have been evaluated using real-world
  process execution logs. However, these evaluations are usually still based on laboratory settings, i.e., the techniques have not yet been applied in a live real-world setting.
\end{description}

The \textbf{treatment design} yields TIDATE as a lightweight process execution and mining framework to
achieve the previously defined goals. For \textbf{treatment validation} TIDATE is validated performing the \textbf{Empirical} cycle of the TAR study.
In the protocol for the TAR study the research problem is defined as follows
using the checklist from \cite{wieringa2014design}.

\begin{description}
  \item[\textbf{Conceptual Framework:}] TIDATE enables the logging of
  process instance executions and the detection of deviating behavior in the execution.
  \item[\textbf{Knowledge Questions:}] \
  \begin{itemize}
    \item How can domain experts easily use process mining techniques?
    \item Can the results provided by process mining techniques be used by domain experts?
  \end{itemize}
  \item[\textbf{Population:}] The population for this approach consists of companies which already have knowledge
  of their processes and focus on the correct execution of process instances.
\end{description}

The validation of TIDATE and the answer for the research problem are
generated by implementing the treatment at a potential client.

For the client selection process, a suitable client was identified through
previous collaborations in different projects. The resources of the client are
reachable through web services. Hence only small adaptations of TIDATE were necessary to deploy it at the client's site.
There are threats to
generalizability, since the research applied the elaborated approach, but
domain experts have been shown the results and developed process models inside
the process execution engine to witness the usability of the approach.

\subsection{Research Design}
\label{sub:design}

After the research context and research problem are defined and a suitable
client is found, the empirical cycle is further defined by the researcher, and
the helper's client cycle (cf. Fig. \ref{fig:tar}) is defined by the researcher
and the client. Both cycles interact with each other, therefore the
coordination starts as soon as the client is acquired and ends after the client
cycle is evaluated (cf. \cite{wieringa2014design}). The problem investigation
is the starting point of this design and adaptation of TIDATE to the client's
needs.

\textbf{Problem Investigation}: The client is interested to know how tasks
are being executed, and if a process model is available, to know if the
execution of instances are matching the behavior of the related model. Common
phenomena for this problem are a different execution behavior due to a missing
automatic task enactment like a process execution engine and missing data sets
allowing the detection of said behavior.

\textbf{Artifact Design}: The artifact TIDATE, that is introduced and implemented at the client, contains a process
execution engine. The client models the manufacturing processes which are then enacted and executed by the process execution engine. Moreover, the engine generates an event stream for the executed process
instances. This event stream is then used for conformance checking to give the
client feedback. Process discovery is not used, since process models have to be created
for the use of the process execution engine. The engine then
orchestrates the execution of active tasks. Hence, no unknown process models
should be discovered from the event stream.

\begin{figure}
  \centering
  \includegraphics[width=0.9\textwidth]{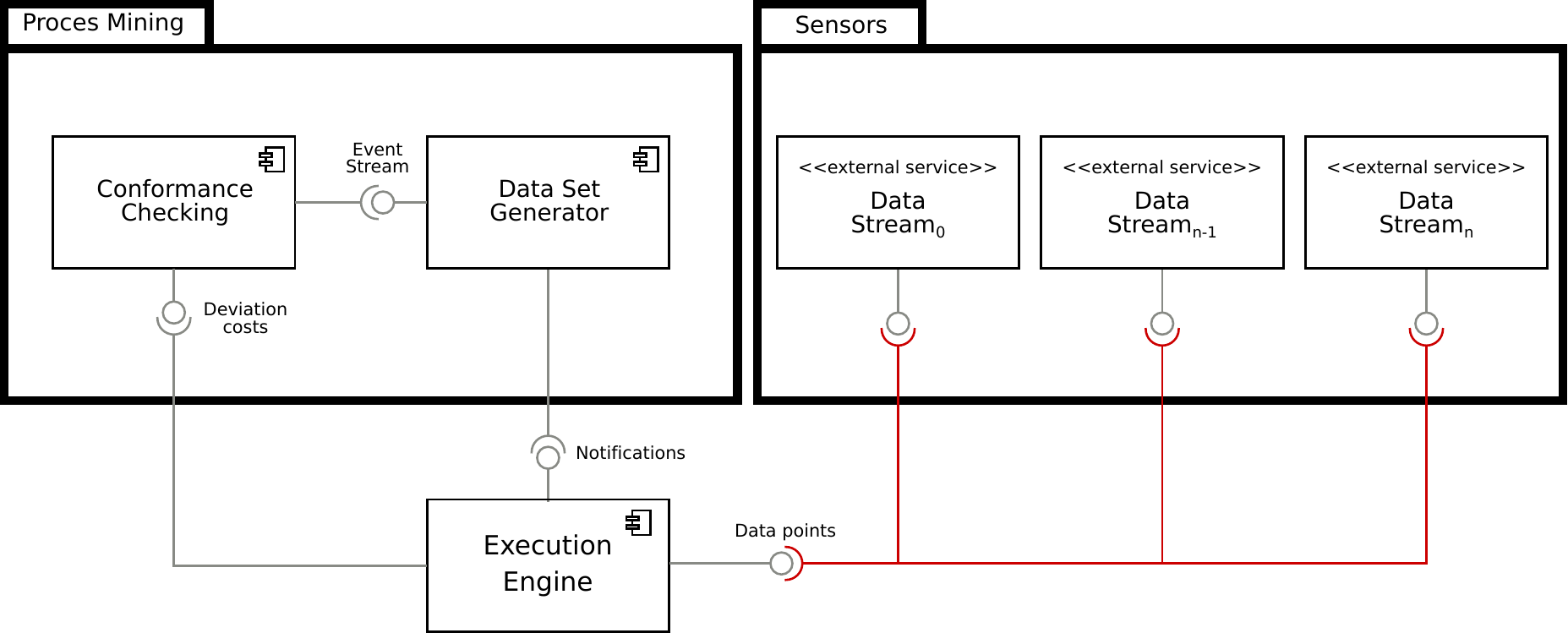}

  \caption{TIDATE architecture}
  \label{fig:arch}
\end{figure}

Fig. \ref{fig:arch} shows the TIDATE architecture. This architecture can be used as blueprint for (lightweigt) process execution and mining frameworks.
The process execution engine creates process instances using process models
from a repository.  The models in the repository are designed by domain
experts. The engine then enacts the tasks in the sequence dictated by the
model. To generate an event stream and a process execution log, a notification
is generated each time a task is enacted, finished, or changed. A description of
the process model is also sent as a notification when a process instance is
created or the model is changed for the process instance.  The Logger component
is subscribed to the engine and generates an event stream based on the
information received from the notifications, creates an process execution log
and stores the description of the related model. This is important for
checking the conformance of an instance.

The conformance checking component is receiving the event stream and the model
description from the Logger component and computes the alignment costs every
time a new event for a trace is detected. Note that the main focus of this
framework is on conformance checking in real time. Since conformance checking of
the workflow perspective is a time consuming task to detect the alignment with
the smallest cost, the framework focuses on conformance of the data elements
. The conformance checking
component is receiving data from external sensors as well. The process
execution engine typically controls the workflow perspective of a process
instance, but the data elements attached to events could still contain wrong
information. In addition, not every information affecting the execution of
a task is shown in an event.  External sources, like the temperature in room
for example, can affect an instance, as well. To exploit this information in conformance
checking, the data from external sensors can be used, which is typically stored
as a time sequence. Time sequences can be compared using, for example, dynamic time warping
\cite{DBLP:conf/bpm/StertzRM20,rakthanmanon2012searching}.

\subsection{Artifact Validation}
\label{sub:validation}

To solve the client's problem of ensuring a correct execution order of the
tasks in a process instance and detecting errors as soon as possible, the
following data is measured. Domain experts design the process model and
together with the researchers create the necessary interfaces for the process
execution engine to interact with the machines. After an introduction, the
domain experts create and start the execution of new process instances. The
event stream is generated by the process execution engine without needed
interference from anyone and the behavior is automatically checked on the data
perspective of the events. This data involves time sequences, temporal
deviations between events and the duration of events as well as other numerical
data relating to the configuration of the participating machines. The
conformance costs are reported back to the process execution engine where
notification events can be produced to inform domain experts. Domain experts
are interviewed in the end to see if process instances that have gone wrong are
detected correctly with an increased cost, if the information was useful in
detecting the exact problem of the process instance as well if problems
occurred during the usage of the framework.

In the following Section \ref{sec:eval}, the artifact is executed inside the client's environment,
results are generated and possible improvements are being created, based on the
feedback of the client.

%% file: eval.tex
In Sect. \ref{sec:meth}, the methodology and planned actions for applying and
executing TIDATE at a client, i.e., a manufacturing environment, were outlined.
This section describes the actual research execution along its setup in Sect.
\ref{sub:setup} and its execution and results in Sect. \ref{sub:execution}.

\subsection{Setup}
\label{sub:setup}

TIDATE as experimental artifact (cf. architecture in Fig. \ref{fig:arch}) is now
implemented at the client. The process execution engine provided by TIDATE, is
used at the client, as it already provides a notification stream to detect
process model changes as well as the enactment and completion of tasks. For
this application, robotic machines and other software interfaces are
orchestrated and controlled by the process execution engine.

\begin{figure}
  \centering
  \includegraphics[width=1.0\textwidth]{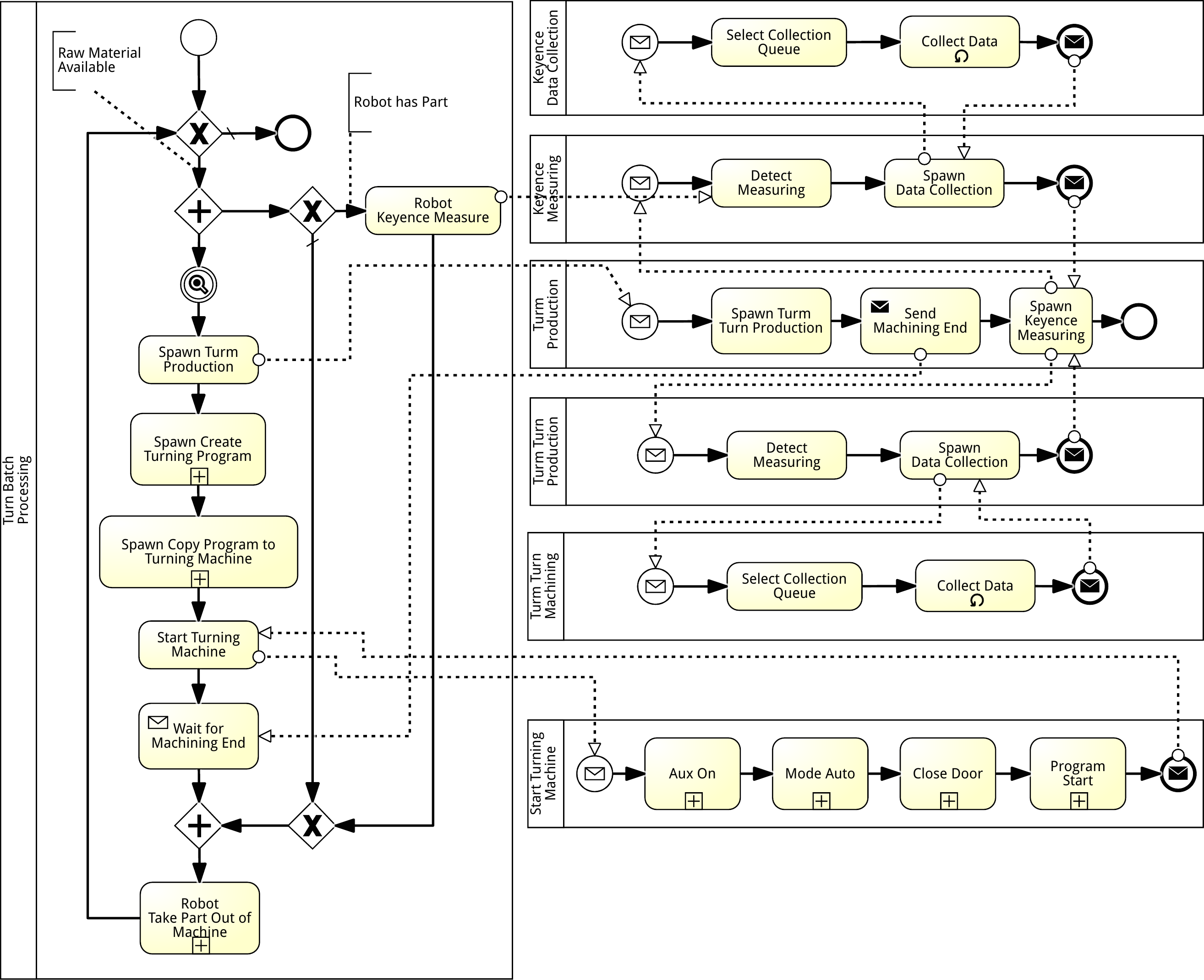}

  \caption{Process model used in the process execution engine for producing \texttt{Turm} parts.}
  \label{fig:megaex}
\end{figure}

A client process driving the production of small metal parts called ``Turm'' by machines for the usage by another company
is depicted in Fig. \ref{fig:megaex}.
In the beginning the machine's state is checked and if it is free, the
production process is spawned. Note that the whole production process is split
into a number of sub processes, so it is easier to read for the domain experts as
well as it is easier to maintain small processes. While different parameters
are fetched during the ``Turn Production'' process, the machine is set up
accordingly with a program to be deployed on the machine. This allows for a high
flexibility to change the process for the machine for each process instance.
When the machine is finished, the part is taken out and measured by the Keyence software
and afterwards manually by domain experts using MicroVU. In the end it, the produced part is put onto a tray outside
of the machine to be used in another process.

The notifications detected by the process execution engine are transformed into an event using the
XES format. These events can then be immediately used by the conformance checking
algorithms. Additional information which is captured by external sensors is
detected by the Fetch task from the ``Turm Keyence Measurement'' process in Fig.
\ref{fig:megaex}. The machine is constantly measuring the diameter of the
produced parts, to check if the parts have been correctly produced.  The
gathered data points of these sensors are aggregated by the software of the
machine, fetched by the process execution engine in the Fetch task and then put into the
notification of the task execution. Hence a time sequence for the diameter of
parts is generated for every process instance. For detecting imperfect parts,
the time sequence of a well produced part is saved additionally in the process execution engine
which is compared by using dynamic time warping \cite{DBLP:conf/bpm/StertzRM20}.

\begin{figure}
  \centering
  \includegraphics[width=0.55\textwidth]{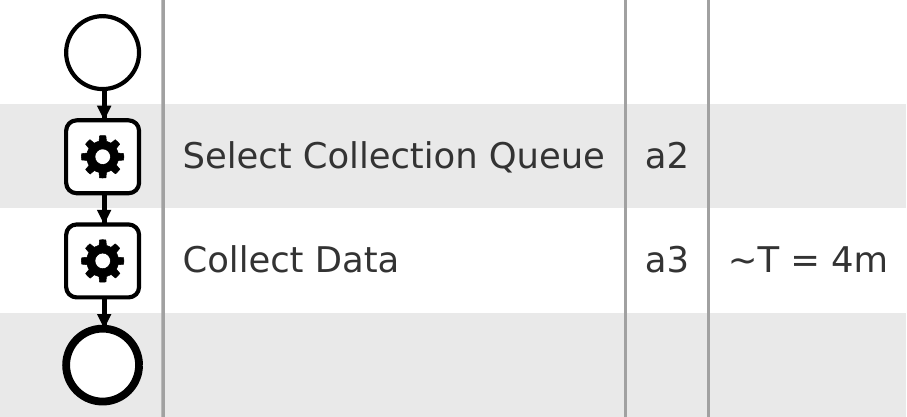}

  \caption{Small example of process model in execution engine at client. The time $\sim{}T$ represents the expected task duration on average.}
  \label{fig:drake}
 \end{figure}

For the detection of temporal deviations, algorithms from
\cite{stertztemp} are used.  There are 2 types of temporal deviations.
The first one is concerning the task duration. If the start and end of a task
is supported by the event stream, the task duration can be calculated by
determining the difference. The other type of deviation that can occur, is the
time distance between the end of a task and the start of the next task. Fig.
\ref{fig:drake} shows a sub-process as implemented in the process execution
engine. For task a3, ``Collect Data'', a time duration is
put into the model. This duration symbolizes the expected task duration on
average. If no time is present, a deviation in this task is seen as harmless.
The \texttt{z-score} \cite{crocker1986introduction} is used to determine the
distance of a witnessed task duration to the task duration in the process
model. If the distance exceeds a threshold, typically 3, an outlier, hence a
severe deviation, is detected.

To apply process mining techniques with correct results, every produced part
should be treated like an process instance. An interesting thing emerged at the
beginning of the implementation and setup of the framework at the client.
Instead of viewing each produced part as a process instance, the shopfloor
workers view the production of a batch as one instance. This means, that if out
of a batch of 5 for example, only one instance shows deviating behavior, the
conformance score is still quite high, since the other 4 instances show no
deviations. For the beginning of this study, the produced parts have been divided hard coded, with
a better solution presented in Sect. \ref{sec:knowledge}.

\subsection{Execution and Results}
\label{sub:execution}

The process execution engine enacts the different tasks for process instances and generates an
event stream. How event streams are produced has been discussed in prior work
\cite{DBLP:conf/bpm/StertzRM20}. This work focuses on discussing the results of the experiment
after several weeks of testing. The data sets for this analysis can be found
here \url{https://tinyurl.com/y6bchfhq/TIDATE/dataset.bise.zip}.

\textbf{Logging}: The client is now able to
automatically generate an event stream and even process execution logs for the
analysis of executed process instances.  This allows for repeatable results
using process mining techniques, since the process execution log contains
time stamps as well to generate an event stream out of it. The logging component
is being executed as a separate process, hence it is never interfering directly
with the execution of process instances.

\textbf{Conformance Checking}: Interesting results emerged from the first batch of parts produced out of the
data set.  By taking a look into the process execution logs of
the different sub processes, no deviations are detected concerning the event
flow of the process instances, but instance \texttt{b20fedd7} of the
``IRB2600 Unload to Tray'' process, contains time deviatons. The ``Move
up'' and ``Move down'' events, which relate to the tasks \texttt{a12} and
\texttt{a21} with a task in between ``wait'' relating to \texttt{a17}. Compared
to other process instances and checked with the domain experts, this
measurement, done by external sensors, takes about $30$ seconds to move a part
through the laser. In the log, all $3$ tasks are taking $0$ seconds. After further
investigation, a software error has been detected, leading to no measurement at
all by the external sensors, which yielded a return value immediately, hence
the task is logged and detected in the event stream as an event, but with an
unexpected time stamp and no values attached to the event. Therefore an error,
has been detected as soon as possible and the external software with sensors
has been restarted to avoid this mistake in future parts.

Another error that has been detected was due to a collision of the robotic
machine arm with another machine. This collision led to a slight misplacement
of a mandrel of the machine. The mandrel is still within a certain safety
range, hence the robotic arm is still reaching inside the machine. However, the
produced part is not at the programmed position due to the misplaced mandrel by
a few millimeter. Hence the robotic arm is still allowed to move, but grabs the
produced part not correctly. The collision is not detectable in the event
stream, since the time stamps, data elements, and event order are correct, but
the time sequence from the external sensors during the measurement of the
current and following parts yields a different time sequence than the expected
time sequence which results in a high distance using dynamic time warping. This
is detected by TIDATE. After investigating the collision, the reason has
been detected. It has been due to a not logged meddling with the program of the
robotic arm, which led to an incorrect execution in the process instance.

Other than the situations described above the process instance of the inspected batch conforms to
the process model and yields correctly produced and placed parts.

The results of the client's cycle are discussed in the following Sect. \ref{sec:knowledge} to conclude
the empirical research cycle.

%% file: knowledge.tex
TIDATE has been implemented, executed, and evaluated at the client. This section discusses how the
previously gathered knowledge questions and improvement goals (cf. Sect. \ref{sub:researchproblem}) can be answered/addressed with the results of the client's cycle.

Overall, the implementation at the client showed that errors in the process execution
can be indeed detected in a real-world setting and help domain experts detect
errors quickly.

\subsection{Knowledge questions}
\label{sub:questions}

The \textbf{first knowledge question} targets the ability to use process
mining techniques easily in such settings.  While the treatment at the client
showed that process mining techniques can be applied easily on event streams
and process execution logs in a real-world setting, the generation of a
sufficient data set needed guidance by researchers. To accomplish the
generation of an event stream using a process execution engine, events have to
be reported back to the engine. Therefore notifications have to be sent to and
from the engine to the services, that are actually performing the tasks. At the
client, this has been done using the HTTP protocol and web services. For tasks
to be executed by human resources, an automatic way to log the execution of an
event is preferable, like in \cite{DBLP:conf/bpm/StertzRM20}.

After setting up a web service that responds to the process execution engine, domain experts by themselves
produced the programs to generate the notifications needed to create an event stream. In addition,
the domain experts independently created new process models and instances without the need of assistance.

Also for the \textbf{second knowledge question}, if domain experts can actually use the
results of process mining techniques, the answer is split into parts.
After a process model is created, the process execution engine ensures the
correct enactment of tasks for a process instance for that model. Standard
conformance checking focuses on the control flow of events, which is a
time consuming task and provides results that should be ensured anyway. The
client is satisfied with the advantages of creating an executable process
model, but is more interested in the data flow of a process instance.  Multi
perspective conformance checking provides the client and the
domain experts with knowledge on process instances going astray through
temporal deviations, analyses the behavior and conformance of data elements
inside and outside an event stream provided through external sensors. These
results are highly appreciated at the client and help domain experts
detecting problems in the execution of an automated process early on. At the moment,
the results of conformance checking are only published as alerts. An implementation
providing a visualization of deviations is desired as an addition to the framework.

\subsection{Improvement goals}
 \label{sub:improvement}
For the \textbf{improvement goals} defined in the research context, the implementation
and execution at the client yielded promising results. An interesting
observation has been made for the generation of suitable data sets. While the
process execution engine is generating an event stream and a log for each
process instance, the fragmentation of the process ``Produce Turm Part'' into
many sub-processes, resulted into events that are assigned to their
individual process instance, but the instance is not related to the production
of a specific part. Therefore all spawned sub-processes by the main processes
are now related to the main process to allow for better results using offline
process mining techniques if desired. Another aspect that we have improved
after the client's cycle, is the separation of produced parts into individual
process instances.  As explained before, every spawned process is related to
the main process in the event stream and log. However, for the interaction with the
process execution engine, all production instances of one batch of parts, are
designed as one process. This process spawns ``Produce Turm Part'' as often as
parts are wanted, i.e., a loop is spawning the production process $5$ times to
produce $5$ parts during one day. To distinguish between a spawned sub-process
and the start of a new main process, we introduced a new icon for BPMN 2.0.

\begin{figure}
  \centering
  \includegraphics[width=\textwidth]{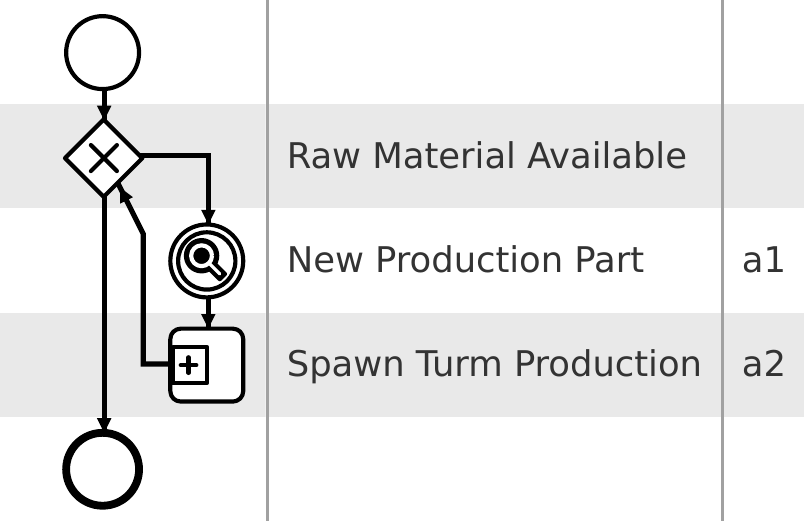}

  \caption{Process model for the daily production of parts. The signal indicates a new online mining item.}
  \label{fig:bpmsignal}
\end{figure}

Fig. \ref{fig:bpmsignal}, shows the process for the production for one day.
The process utilizes a custom BPMN event (intermediate throwing event,
magnifying glass), which allows to group log-data based on individually
produced parts, instead of saving it solely based on the given processes
structure.  While industrial processes are often modeled around the interaction
of machines and the production of batches (restricted by raw material
availability in the machine), the information derived from online mining is
expected to be about individually produced parts. Thus when a process is
modeled from a machine coordination point-of-view, a single instance contains
information about multiple produced parts. The custom BPMN event is a simple
way tell where the event-stream for a single parts, thus allowing to
automatically extract and separate part-information.

The highlighting of deviating data elements is already explained in Sect.
\ref{sec:eval}, where the alert of temporal deviations and the increased
distance between two time sequences, helped detecting incorrectly executed
process instances and discovering the source for the error. So both improvement
goals (cf. Sect. \ref{sub:researchproblem}) have been reached. The last improvement goal of highlighting deviating
behavior in the execution order, cannot be fully answered at this moment.
Since the process execution engine is ensuring the order of
the tasks to be fulfilled, the main reason for deviating behavior is an error
in the logging component or an attacker tampering with the process execution
engine. None of these two scenarios has taken place in the study at the client.
While there are studies for online conformance checking, i.e., by
\cite{van2017online}, using real life data, we could not verify it in our
study, because of not occurring. Since conformance checking for deviating behavior
in the order of events is quite time consuming, this result can still show the benefit
of having a process execution engine enacting process instances in an organization.

For the research goal, we can conclude that an online process mining framework,
indeed supports domain experts in an environment with defined process models.
After the implementation, domain experts can model new process models, start
and execute process instances. Instances containing errors in the execution are
reported to the experts and offer an explanation for incorrect products.  The
implementation at the client yielded promising results. There are some threats
to generalizability, since the manufacturing domain offers process models,
which are usually already known by the organization and focuses on conformance
checking. Other domains with increased human interactions which do not follow a
strict structure could yield incorrect results. Further studies in different
domains are required.

%% file: rel.tex
A plethora of process mining techniques has emerged over the last years, dealing
with the three areas of process mining: process discovery, conformance checking
and, process enhancement \cite{vanderAalst:2011:PMD:1983446,berti2019process}. Existing case
studies, e.g., \cite{badakhshan2020using,dakic2019process} apply all three process mining
ares to real-world
process execution logs, using existing tools such as Disco
\cite{gunther2012disco} or ProM \cite{van2005prom}. By contrast to this study,
these approaches focus on the applicability of process discovery on real-world logs in an offline manner and the associated filtering of the event log,
since without a process execution engine, some noise is expected in the event
logs. Conformance checking is evaluated as well, but on the order of events, as
well as some social network mining of the resources and various improvements
for the process models focusing on the time perspective with process
enhancement. 

The usage of process mining in organizations and how to start an
enterprise with process mining in mind, is explained in \cite{Rein2020}. Here
several best practices are presented from projects in different organizations,
like Siemens. BMW and Uber. The main difference to the TIDATE framework is that
the best practice use cases aim at discovering complex process models, which
are difficult to understand and design. Therefore the process model is created
retrospectively using process discovery on a process execution log. This is due
the structure of the organizations and the interaction with other organizations
involving human resources. By contrast, in this study, the
process models are designed by domain experts to be executed by a process execution engine
and the conformance of the process instance is monitored with focus on data and time during runtime. 

The work of \cite{leemans2019directly}, argues that even though process
discovery is commonly present in commercial tools nowadays, the discovered
process models are often only visualized in a ``directly-follows'' graph
representation. This simple representation can lead to false conclusions and
assumptions about the process model, since it is not clear, where parallel
activities or decisions are taking place and what leads to a specific path in
the process model. Hence, \cite{leemans2019directly} introduces an extension to ``directly-follows'' models.
Moreover, the approach shows how to apply conformance checking based on the extended models.
The approach is evaluated in a real-world
setting. In detail, conformance checking is conducted by comparing process
models, created by hand, to the generated process execution log. Contrary to
the work in this paper, the order of the events is the important aspect when conducting conformance
checking, not the focus on data and time.

%% file: concl.tex
This paper demonstrates how to design and use a process execution and mining framework in manufacturing.
In particular, the experimental framework TIDATE
has been implemented in a client's manufacturing environment following a Technical Action
Research study. 

The implementation at the client
\begin{itemize}
\item revealed differences between a
real-world setting and the academic environment, in particular, the
different perspectives on process instances, i.e., batch-oriented vs. product-oriented.
\item led to answers to research and improvement questions, for example, how process mining can be integrated in the manufacturing domain and if domain experts can use these techniques and with how much effort
required. 
\end{itemize}

The introduction of a process execution engine and the nature of the
manufacturing domain, helped applying process mining and the results look
promising. Time deviations and deviations in data elements, even from
external sources, like data sensors, are providing much needed information on
the conformance of a process instance and give a close inspection to analyze
where a process instance is deviating from the expected behavior.

Since not every domain has the same degree of process orientation as the
manufacturing domain, conducting a TAR study with TIDATE in another domain is desired.
First of all, such a study could shed light on the potential of process discovery
and conformance checking on the order of events, since this only played a
minor role in this study. Moreover, domains such as health care and logistics seem to have
similar demands with respect to time and data-aware conformance checking. In both domains, sensors play an
important role and time is a critical factor. 

Another important aspect for future work is how the information on deviations is conveyed to the
clients, i.e., domain experts. In this study, simple alert notifications are
sent if a deviation is detected, but a visualization of the deviating tasks of
the process model in the process execution engine, would lower the entry
barrier even more for domain experts to get accustomed to process mining.

%% file: main.bbl
\begin{thebibliography}{10}
\providecommand{\url}[1]{\texttt{#1}}
\providecommand{\urlprefix}{URL }
\providecommand{\doi}[1]{https://doi.org/#1}

\bibitem{vanderAalst:2011:PMD:1983446}
van~der Aalst, W.M.P.: Process Mining: Discovery, Conformance and Enhancement
  of Business Processes. Springer (2011)

\bibitem{badakhshan2020using}
Badakhshan, P., Alibabaei, A.: Using process mining for process analysis
  improvement in pre-hospital emergency. In: ICT for an Inclusive World, pp.
  567--580. Springer (2020)

\bibitem{berti2019process}
Berti, A., van Zelst, S.J., van~der Aalst, W.: Process mining for python
  (pm4py): bridging the gap between process-and data science. arXiv preprint
  arXiv:1905.06169  (2019)

\bibitem{DBLP:books/sp/CarmonaDSW18}
Carmona, J., van Dongen, B.F., Solti, A., Weidlich, M.: Conformance Checking -
  Relating Processes and Models. Springer (2018).
  \doi{10.1007/978-3-319-99414-7},
  \url{https://doi.org/10.1007/978-3-319-99414-7}

\bibitem{crocker1986introduction}
Crocker, L., Algina, J.: Introduction to classical and modern test theory. ERIC
  (1986)

\bibitem{dakic2019process}
Dakic, D., Sladojevic, S., Lolic, T., Stefanovic, D.: Process mining
  possibilities and challenges: A case study. In: 2019 IEEE 17th International
  Symposium on Intelligent Systems and Informatics (SISY). pp. 000161--000166.
  IEEE (2019)

\bibitem{gunther2012disco}
G{\"u}nther, C.W., Rozinat, A.: Disco: Discover your processes. BPM (Demos)
  \textbf{940},  40--44 (2012)

\bibitem{Gartner2020}
Kerremans, M., Searle, S., Srivastava, T., Iijima, K.: Market guide for process
  mining (2020), \url{www.gartner.com}

\bibitem{leemans2019directly}
Leemans, S.J., Poppe, E., Wynn, M.T.: Directly follows-based process mining:
  Exploration \& a case study. In: 2019 International Conference on Process
  Mining (ICPM). pp. 25--32. IEEE (2019)

\bibitem{rakthanmanon2012searching}
Rakthanmanon, T., Campana, B., Mueen, A., Batista, G., Westover, B., Zhu, Q.,
  Zakaria, J., Keogh, E.: Searching and mining trillions of time series
  subsequences under dynamic time warping. In: Proceedings of the 18th ACM
  SIGKDD international conference on Knowledge discovery and data mining. pp.
  262--270 (2012)

\bibitem{Rein2020}
Reinkemeyer, L.: Process Mining in Action -- Principles, Use Cases and Outlook.
  Springer International Publishing (2020). \doi{10.1007/978-3-030-40172-6}

\bibitem{stertztemp}
Stertz, F., Mangler, J., Rinderle{-}Ma, S.: Temporal conformance checking at
  runtime based on time-infused process models. CoRR  \textbf{abs/2008.07262}
  (2020), \url{https://arxiv.org/abs/2008.07262}

\bibitem{DBLP:conf/bpm/StertzRM20}
Stertz, F., Rinderle{-}Ma, S., Mangler, J.: Analyzing process concept drifts
  based on sensor event streams during runtime. In: Business Process
  Management. pp. 202--219 (2020). \doi{10.1007/978-3-030-58666-9\_12}

\bibitem{van2005prom}
Van~Dongen, B.F., de~Medeiros, A.K.A., Verbeek, H., Weijters, A., van
  Der~Aalst, W.M.: The prom framework: A new era in process mining tool
  support. In: International conference on application and theory of petri
  nets. pp. 444--454. Springer (2005)

\bibitem{wieringa2012technical}
Wieringa, R., Moral{\i}, A.: Technical action research as a validation method
  in information systems design science. In: International Conference on Design
  Science Research in Information Systems. pp. 220--238. Springer (2012)

\bibitem{wieringa2014design}
Wieringa, R.J.: Design science methodology for information systems and software
  engineering. Springer (2014)

\bibitem{van2017online}
van Zelst, S.J., Bolt, A., Hassani, M., van Dongen, B.F., van~der Aalst, W.M.:
  Online conformance checking: relating event streams to process models using
  prefix-alignments. Data Science and Analytics pp. 1--16 (2017)

\end{thebibliography}
